\documentclass[11pt]{article}

\usepackage[final]{acl}

\usepackage{times}
\usepackage{latexsym}
\usepackage[T1]{fontenc}
\usepackage[utf8]{inputenc}
\usepackage{microtype}
\usepackage{inconsolata}
\usepackage{xurl}
\usepackage{graphicx}
\IfFileExists{latexml.sty}{\usepackage{latexml}}{\newif\iflatexml}
\usepackage{booktabs}
\usepackage{amsmath}
\usepackage{multirow}
\usepackage{xcolor}
\usepackage{enumitem}

\newcommand{\xmult}[1]{\iflatexml #1\texttimes\else\ensuremath{#1\times}\fi}
\newcommand{\numlist}[1]{\iflatexml #1\else\ensuremath{#1}\fi}

\title{Enrich-Retrieve-Rank: Scaling Capability Discovery\\
       Beyond In-Context Routing}

\iflatexml
  \author{{Nazib Sorathiya, Daniel Zhang, Bardiya Akhbari} \\
    Amazon AGI \\
    Seattle, WA, USA \\
    \texttt{\{nsorath, nyudan, bardiyaa\}@amazon.com}}
\else
  \author{Nazib Sorathiya \quad Daniel Zhang \quad Bardiya Akhbari \\
    Amazon AGI \\
    Seattle, WA, USA \\
    \texttt{\{nsorath, nyudan, bardiyaa\}@amazon.com}}
\fi

\date{}

\begin{document}
\maketitle

\begin{abstract}
Agent ecosystems now include thousands of \textbf{MATS} components (\textbf{M}odels, \textbf{A}gents, \textbf{T}ools, and \textbf{S}kills), yet their discovery still relies on in-context routing. These systems read a registry (names, hints, or descriptions, as context budget permits), pick a candidate, invoke it, and retry on failure. This pattern degrades with scale, and registries are growing fast. We recast capability discovery as search over a registry by defining an offline enrichment step that turns sparse metadata into searchable profiles, and an online retrieve-then-rank pipeline that returns a ranked shortlist \emph{without invoking} any candidates online. We show that from $N=10$ to $7{,}278$ capabilities, in-context routing's top-1 accuracy (Match@1) collapses ($0.85 \to 0.12$), while retrieve-then-rank degrades more gently ($0.81 \to 0.39$) because its reranker still ranks the right capability first $0.70$--$0.87$ of the time once retrieval finds it. In the Nova Micro sweep, the crossover is around $N=500$. We compare against two in-context baselines. Full-Ctx puts the whole registry in the prompt and asks the LLM to pick. Search\&Pick gives the LLM a search tool to narrow candidates before it picks. At full scale the pipeline leads Search\&Pick by 6.5 percentage points (pp) on Match@1 at about half the cost, and reduces the cost \xmult{70} versus Full-Ctx. We use the same configuration (same enrichment, retriever, and scorer weights) across agent, tool, and skill registries. The pipeline runs in production as the default capability-discovery layer of a large-scale multi-agent platform.
\end{abstract}

\section{Introduction}
\label{sec:intro}

\begin{figure*}[t]
\centering
\iflatexml
  \includegraphics[width=\textwidth]{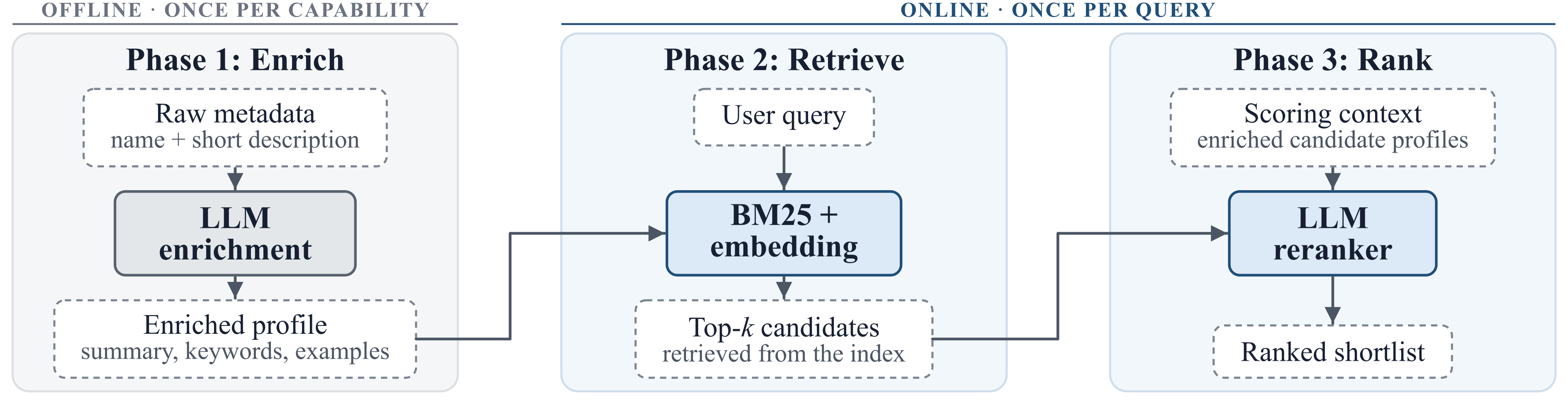}
\else
  \includegraphics[width=\textwidth]{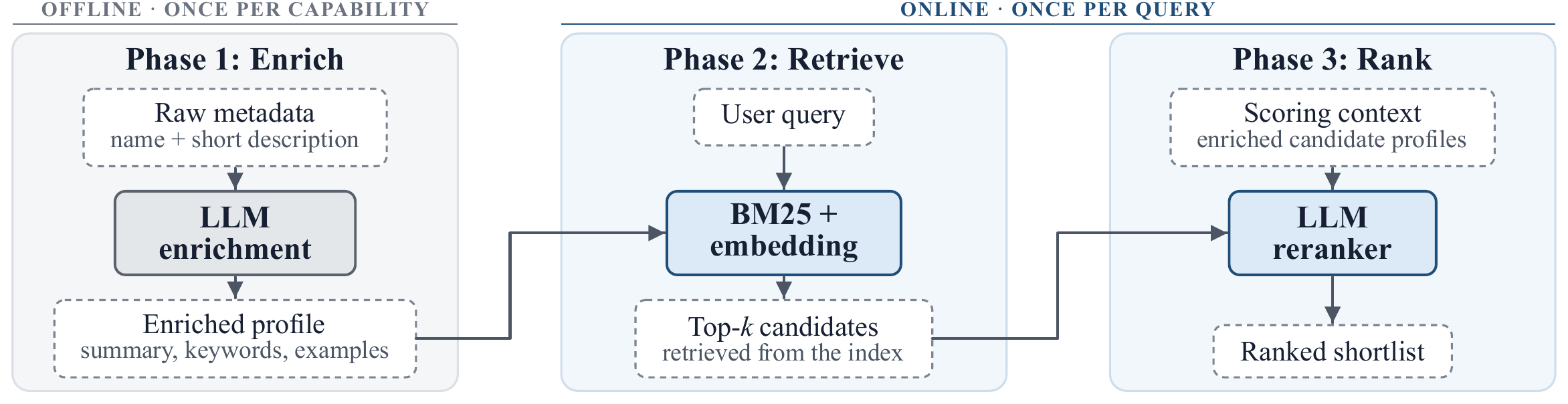}
\fi
\caption{The Enrich-Retrieve-Rank pipeline. Phase~1 transforms sparse capability metadata into rich profiles. Phase~2 retrieves top-$k$ candidates from the enriched index. Phase~3 ranks with a single LLM call. The entire online path requires one LLM call at bounded latency. Dashed boxes are data, and solid boxes are computation.}
\label{fig:pipeline}
\end{figure*}

Recent agentic systems \citep{claudecode,cursor,langgraph,bedrockagents} expose registries of thousands of components. Our production registry unifies all MATS components under one schema and already exceeds $500$ entries in every MATS type; the largest benchmark we study holds $7{,}278$ tools. At that scale an orchestrator cannot place every candidate in context, so choosing what to call requires first narrowing what to consider.

Today's systems discover capabilities by in-context routing. The large language model (LLM) sees a representation of the registry (tool names, one-line hints, or full descriptions as budget permits), picks a candidate, invokes it, and retries on failure. Some systems truncate alphabetically; others use a regular expression tool to narrow candidates before picking. All couple cost and accuracy to registry size. Every wrong invocation costs tokens and latency, and may call an untrusted endpoint just to learn what it does.

We frame capability discovery as web search over a capability registry rather than the open web. A search engine does not make users click through every indexed page, and an orchestrator should not present every capability and rely on the LLM to pick. We retrieve ranked candidates \emph{without invoking them} and return a shortlist. The advantage over in-context routing \emph{widens with registry size}, so it matters most where the field is heading.

We decompose end-to-end performance into retrieval recall and reranker conditional accuracy. The reranker reaches near-peak accuracy at modest registry sizes and stays flat beyond that, so at larger scales most misses originate in retrieval. At full scale, our pipeline leads Search\&Pick in match accuracy and uses half the tokens.

We apply one pipeline to agent, tool, and skill registries under a single configuration and deploy it in production as the capability-discovery layer for a multi-agent platform. The pipeline includes offline enrichment (an LLM rewrites sparse metadata into rich profiles once at registration time, not per query). Its value scales with metadata sparsity. On well-documented public data the effect is neutral-to-negative (\S\ref{sec:exp-ablations}). We include it as a production design choice, not a source of benchmark gains.

\section{Related work}
\label{sec:related}

\paragraph{Tool retrieval.}
Toolformer \citep{schick2023toolformer} self-supervises a fixed API set; Gorilla \citep{patil2023gorilla} trains end-to-end on APIs; ToolLLM \citep{qin2024toolllm} scales the endpoints with a retriever and depth-first-search caller.
All fix the API pool at train time, leaving retrieval implicit.
ToolRet \citep{shi2025toolret} strips the pre-annotated tool list and shows that general information retrieval (IR) models perform poorly on the resulting task.
CRAFT \citep{yuan2024craft} and EasyTool \citep{yuan2024easytool} are the closest enrichment-side methods. CRAFT abstracts code into per-task toolsets; EasyTool rewrites API docs into instructions.
PLUTO \citep{huang2024pluto} pairs planning with description editing and is the closest prior recipe to ours. It does not quantify the retrieve-vs-rerank decomposition or study scale.
AnyTool \citep{du2024anytool} uses a hierarchical agent over 16k APIs but does not separate retrieval from MATS invocation.
MetaTool \citep{huang2024metatool} and \citet{qu2024toolsurvey} confirm tool selection is the dominant failure mode but propose no retrieve-then-rank pipeline.

\paragraph{Agent routing.}
AppWorld \citep{trivedi2024appworld} keeps agent selection implicit because the registry fits in context ($n=8$).
HuggingGPT \citep{shen2023hugginggpt} and Visual ChatGPT \citep{wu2023visualchatgpt} pick from a static menu. Our Nova Micro sweep shows that this pattern breaks around $N=500$.
AFlow \citep{zhang2024aflow} and \citet{guo2024maissurvey} frame the challenge as workflow construction over heterogeneous agents; we contribute the retrieval primitive.

\paragraph{Capability discovery.}
The Model Context Protocol (MCP) \citep{mcp2024}, Agent2Agent \citep{a2a2025}, and the GPT Store \citep{gptstore2024} standardize capability enumeration and invocation but assume the client already knows \emph{which} server to call.
ToolkenGPT \citep{hao2023toolkengpt} embeds tools as soft tokens but ties embeddings to a specific checkpoint.
\citet{mulang2026kgtools} instead use a knowledge-graph index over MCP servers; their graph-based filtering improves discovery at 269 tools but not on smaller curated menus.

\vspace{-0.4em}
\paragraph{IR, dense embeddings, and LLM reranking.}
Capability discovery mirrors web search as the system matches a user intent against a large corpus of documents (capabilities) indexed offline.
Our pipeline reuses the retrieve-then-rerank architecture web search engines refined over decades, adapted to capability metadata rather than web pages.
BM25 \citep{robertson2009bm25} is our lexical baseline.
doc2query \citep{nogueira2019doc2query} expands documents at index time with predicted queries; our enrichment writes typed structured fields instead, consumed by both retriever and reranker.
HyDE \citep{gao2022hyde} generates a hypothetical document at query time; we do the analogous work once at registration.
For dense retrieval we use BGE-large-en-v1.5 \citep{xiao2024cpack} and Amazon Titan Embed V2 (1024d); E5-Mistral \citep{wang2024e5mistral} and GTE \citep{li2023gte} are left to future work.
RankGPT \citep{sun2023rankgpt} showed listwise LLM rerankers match supervised cross-encoders.
Two concurrent works apply enrich-then-retrieve to tools independently: Tool-DE \citep{lu2025toolde} and Multi-Field Tool Retrieval \citep{tang2026mftr}.
Neither studies how performance changes as the registry grows.

\vspace{-0.3em}
\paragraph{Our contribution.}
This is a systems-and-scaling study, not a new-model study.
The pipeline composes established components (offline enrichment \citep{gao2022hyde,nogueira2019doc2query,yuan2024craft,yuan2024easytool,huang2024pluto}; BM25/BGE retrieval; listwise reranking \citep{sun2023rankgpt}); three contributions are new.
First, we map the full degradation curve from $N=10$ to $7{,}278$ and locate the crossover around $N=500$ in the Nova Micro sweep. Prior work reports degradation at a single scale \citep{shi2025toolret,du2024anytool}.
Second, we decompose retrieval from reranking. The fixed-$k$ reranker holds a $0.70$--$0.87$ conditional-accuracy band, while ${\sim}70\%$ of large-registry \emph{misses} occur at retrieval.
Third, we compare our results against trial-and-error baselines grounded in how deployed harnesses route, measuring accuracy, tokens, latency, and cost (absent from prior literature).

\section{Method}
\label{sec:method}

\subsection{Problem formulation}
\label{sec:method-problem}
Given a query $q$ and a capability registry $C = \{c_1, \dots, c_N\}$, we produce a ranked list $\{(c, s_c)\}$ without invoking any $c_i$ online. Each $c_i$ belongs to one type; our experiments cover Tools, Agents, and Skills, leaving model routing to existing work.
The orchestrator consumes the top-$k$ list and picks.

\subsection{Offline enrichment}
\label{sec:method-enrich}
At registration time (once per capability, never per query), an LLM rewrites sparse metadata into a structured profile with five typed fields of a capability summary, an action-verb-led description, differentiating keywords, and positive and negative usage examples.
We add a trust score and self-reported capability tags for production registries.

These typed fields feed \emph{both} the retriever and the reranker, unlike HyDE \citep{gao2022hyde}, which generates a hypothetical document per query at inference time, or doc2query \citep{nogueira2019doc2query}, which expands documents with surrogate queries for a single retrieval index.
The retriever indexes them lexically and densely; the reranker reads them verbatim when scoring candidates.
We concatenate these fields into the BM25 index and encode with BGE-large-en-v1.5 or Amazon Titan Embed V2.

\subsection{Online retrieve-then-rank}
\label{sec:method-online}
We retrieve the top $k \in \{15, 25\}$ candidates using BM25, neural, or hybrid retrieval. We then combine four scores into $[0,1]$: LLM (weight 0.50, one API call), BM25 (0.05), Quality (0.30), and Intent (0.15). Quality and Intent drop out on public registries because they lack trust and type metadata.

The two production-only signals use fields from the enriched profile (\S\ref{sec:method-enrich}). \emph{Quality} is the normalized offline trust score; \emph{Intent} matches a query-inferred capability type to the candidate's self-reported type tags. A signal drops out when its fields are missing, and the remaining weights are renormalized. Thus, all public results use only LLM and BM25 at a fixed $10{:}1$ ratio. We set the four weights once according to each signal's role and do not tune them by capability type or dataset.

\section{Experiments}
\label{sec:exp}
\subsection{Setup}
\label{sec:exp-setup}

We evaluate our pipeline across three capability types and five dataset/type combinations (Table~\ref{tab:setup}). We report \textbf{Match@$k$} ($k \in \{1,3,5\}$) as the fraction of queries for which any ground-truth capability appears in the top-$k$ returned list, \textbf{mean reciprocal rank (MRR)} as the average of $1/\mathrm{rank}$ of the first correct result, and \textbf{Recall@$k$} as the fraction of ground-truth capabilities retrieved in the top-$k$ (relevant when a query maps to multiple ground-truth).
We compute bootstrap 95\% confidence intervals (CIs) via $n=1000$ resamples; all reported deltas $\geq 5$~pp exceed the 95\% CI width ($\pm1.5$--$3$~pp on ToolRet, $\pm5$--$8$~pp on the smaller AppWorld data).

Unless noted, we use Nova Micro as the reranker. A four-model experiment (Nova~Micro, Nova~Lite, Claude~3.5~Haiku, Claude~Sonnet~4) confirms the method ordering is stable across rerankers.

\begin{table}[ht]
\centering
\footnotesize
\setlength{\tabcolsep}{5pt}
\begin{tabular}{llrr}
\toprule
Type & Benchmark & Reg. & Queries \\
\midrule
Tools  & ToolRet    & 7{,}278 & 7{,}961 \\
Tools  & AppWorld   & 332     & 147     \\
Agents & ToolRet    & 5{,}885 & 7{,}961 \\
Agents & AppWorld   & 8       & 147   \\
Skills & MCP        & 859     & 1{,}627 \\
\bottomrule
\end{tabular}
\caption{Capability $\times$ benchmark coverage.}
\vspace{-0.5em}
\label{tab:setup}
\end{table}

\vspace{-0.9em}
\subsection{Results}
\label{sec:exp-main}

We report Match@1 across the benchmarks and methods (Table~\ref{tab:main-mat1}). Each baseline mirrors a routing strategy used in production: Regex (pattern-match, no LLM); Full-Ctx (full registry in-context, LLM picks); Search\&Pick (LLM with a search tool to narrow candidates before picking); Trial\&Err (LLM iteratively lists and tries).\footnote{Trial\&Err is scored on its final try. It can fall below the $1/n$ random-pick rate when it exhausts its round budget without committing. Agents-AppWorld picks an agent on 2 of 147, scoring $0.014$ vs.\ $0.125$.}
We also include two retriever-only baselines (BM25 and BGE-large-en-v1.5), and three full-pipeline configurations (Ours+BM25, Ours+BGE-5f, and Ours+Titan).

We use Match@1 as our primary metric because it allows direct comparison with the single-pick baselines. The pipeline itself returns a ranked shortlist ($k\in\{15,25\}$, with $k=15$ in production), so we also report Match@3, Match@5, MRR, and Recall@15 to measure whether the correct capability reaches a downstream selector. On Tools-ToolRet, for example, Ours+Titan reaches Match@3 $0.523$ and Match@5 $0.558$. Full results are reported in Appendix~\ref{app:full-table1} (Tables~\ref{tab:rank-metrics},~\ref{tab:r15}).

\begin{table*}[t]
\centering
\small
\setlength{\tabcolsep}{4pt}
\begin{tabular}{lccccccccc}
\toprule
Row (\textit{n} / \textit{q}) & Regex & \shortstack{Full\\Ctx} & \shortstack{Search\\\&Pick} & \shortstack{Trial\\\&Err} & BM25 & \shortstack{BGE\\5f} & \shortstack{Ours\\+BM25} & \shortstack{Ours\\+BGE} & \shortstack{Ours\\+Titan} \\
\midrule
Tools / ToolRet (7{,}278 / 7{,}961)  & 0.231 & 0.166 & 0.332 & 0.054 & 0.311 & 0.346 & 0.388 & 0.389 & \textbf{0.397} \\
Tools / AppWorld (332 / 147) & 0.408 & \textbf{0.762} & 0.476 & 0.061 & 0.483 & 0.429 & 0.741 & 0.735 & 0.741 \\
Agents / ToolRet (5{,}885 / 7{,}961) & 0.228 & 0.148 & 0.341 & 0.045 & 0.336 & 0.360 & \textbf{0.418} & 0.412 & 0.397 \\
Agents / AppWorld (8 / 147)& 0.993 & 0.986 & 0.973 & 0.014 & \textbf{1.000} & \textbf{1.000} & \textbf{1.000} & \textbf{1.000} & \textbf{1.000} \\
Skills / MCP, leaky (859 / 1{,}627) & 0.648 & \textbf{0.968} & 0.829 & 0.624 & 0.880 & 0.892 & 0.807 & 0.806 & 0.942 \\
\bottomrule
\end{tabular}
\caption{Match@1 splits by registry size ($n$) and query count ($q$): the pipeline leads on large registries while in-context routing wins for the smaller sizes. Ours+BGE is the 5-field BGE config. Bold marks the best result in each row. Ours+Titan uses Amazon Titan Embed V2 as the dense retriever. Full-Ctx uses an $n=500$ sub-sample on Tools-ToolRet, Agents-ToolRet, and Skills-MCP because the full registry exceeds Nova Micro's context window; every other cell is full-$n$. Trial\&Err scores its final \texttt{try} and can fall below the $1/n$ random-pick rate when it abstains rather than guessing; see footnote. Other metrics are in
Appendix~\ref{app:full-table1}.}
\vspace{-1.4em}
\label{tab:main-mat1}
\end{table*}

On Tools-ToolRet (7,961 queries, $N$=7,278 capabilities), Ours+Titan reaches Match@1 0.397, a 6.5~pp lead over Search\&Pick (0.332) at half the token cost. The gain is broad as the pipeline beats Search\&Pick on 13/16 Tools and 12/16 Agents sources with $n\geq100$ (Appendix~\ref{app:bysource}). ToolRet includes near-duplicate items, and the error decomposition shows $70\%$ of failures occur at retrieval. On small registries the pipeline loses its advantage. Full-Ctx leads on Tools-AppWorld ($n$=332, 0.76 vs.\ 0.74) and Skills-MCP. The Nova Micro sweep places the crossover near $N=500$.\footnote{Full-Ctx runs on a 500-capability subsample because the full registry exceeds Nova Micro's context window.}

Agents-ToolRet shares queries and 89\% of its registry with Tools-ToolRet, so it is a near-replicate rather than independent evidence. Agents-AppWorld saturates at 1.00 across its eight-agent registry. Skills-MCP's query set paraphrases capability descriptions, giving lexical retrievers an edge. Public benchmarks do not yet support an independent agent- or skill-retrieval win at scale. Even on Skills-MCP, Ours+Titan (0.942) closes most of the gap to Full-Ctx (0.968), up from 0.807 with Ours+BM25. Titan's retrieval recall accounts for the gain (R@15 $0.994$).

\subsection{Ablations}
\label{sec:exp-ablations}

We compare raw vs.\ LLM-enriched registries with everything else held fixed (same query set, registry size, retriever, $k$, weights, model; full numbers in Appendix~\ref{app:detail-tables}, Table~\ref{tab:ablation}).
Enrichment is neutral-to-negative on public data: $\pm0.1$~pp Match@1 on ToolRet, $0.0$~pp on AppWorld (with $+3.4$~pp Match@3), and $-4.4$~pp on MCP where added keywords dilute already-strong lexical overlap.
A per-field ablation (ToolRet $n=500$) shows no single field raises Match@1 by more than $+0.6$~pp (keywords, $0.842 \to 0.848$).
Public registries already have clean, detailed descriptions, so enrichment fills no gap. Its production value scales with metadata sparsity, which public benchmarks cannot stress-test.
We therefore construct a controlled stress test by degrading the same 1{,}000-tool ToolRet pool from full descriptions to first-sentence hints and names only, then comparing raw and enriched registries with the pipeline fixed. Enrichment improves Match@1 by $+5.8$, $+9.1$, and $+25.6$~pp, respectively (paired McNemar $p<10^{-6}$). At name-only, Recall@15 rises from $0.134$ to $0.467$. The end-to-end gain therefore originates in retrieval (Figure~\ref{fig:sparsity}; Appendix~\ref{app:sparsity}).

\subsection{Retrieval analysis}
\label{sec:exp-retrieval}

On large-registry ToolRet, ${\sim}70\%$ of misses occur at retrieval because BM25's top 15 excludes the ground-truth capability (68\% of Tools-ToolRet and 70\% of Agents-ToolRet misses; Cat~1, Appendix~\ref{app:failures}). AppWorld and MCP misses occur during reranking.
Both neural retrievers raise Recall@15 over BM25 by $+4.7$--$7.6$~pp on the ToolRet rows (BGE-5f and Titan Embed V2; full numbers in Appendix~\ref{app:full-table1}). The gain does not propagate cleanly to end-to-end Match@1. The three pipeline configurations sit within $0.9$~pp on Tools-ToolRet (0.388, 0.389, 0.397) and $2.1$~pp on Agents-ToolRet (0.397, 0.412, 0.418) because the fixed-$k$ reranker's conditional accuracy is flat across retrievers.
Titan attains the best end-to-end Match@1 on Tools-ToolRet (0.397); on Agents-ToolRet, BM25 retrieval leads (0.418). We adopt Titan as the production retriever for its stronger recall (R@15 0.625 vs BM25 0.549) and its leading Match@1 on the Tools and Skills rows. The remaining recall gap is real, but the dense encoders close little of it end-to-end. Doing so requires a much stronger first-stage retriever, not a better reranker.
ToolBench-IR, a BERT-base retriever fine-tuned for tool-API relevance \citep{qin2024toolllm}, does not improve the candidate pool. Its Recall@15 is $0.608$ on Tools and $0.596$ on Agents, below the general encoders ($0.625$ and $0.626$); end-to-end Match@1 is lower ($0.378$ vs.\ $0.397$ and $0.381$ vs.\ $0.412$; Appendix~\ref{app:tooltuned}). Existing domain-specific fine-tuning therefore does not remove the first-stage recall bottleneck.

\subsection{Scale and latency}
\label{sec:exp-scale}
\begin{figure*}[t]
\centering
\includegraphics[width=0.9\textwidth]{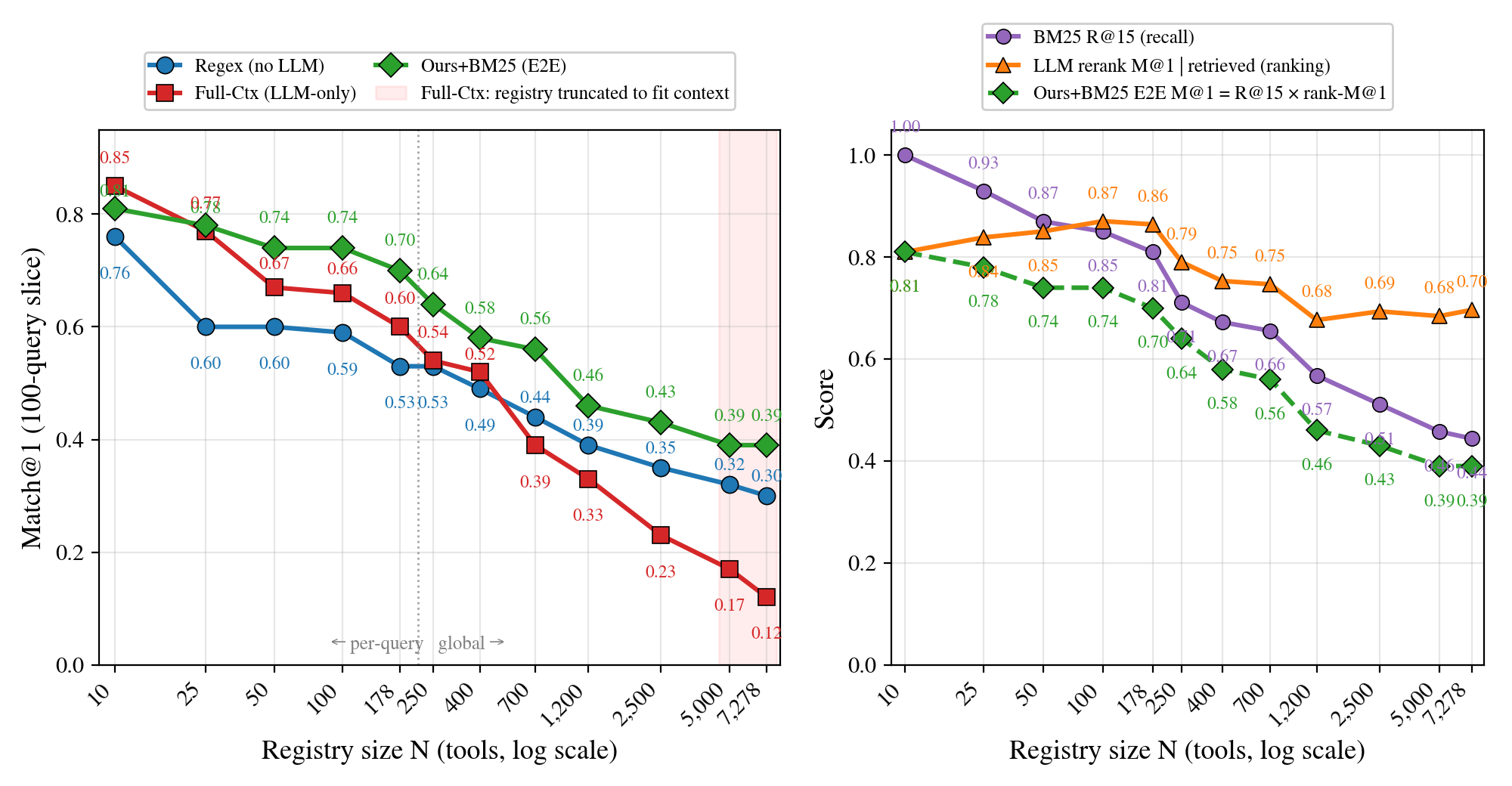}
\caption{Within-benchmark scaling. Match@1 vs.\ $N$ (Left) and pipeline decomposition (Right).}
\vspace{-1em}
\label{fig:scaling-within}
\end{figure*}

We ran a registry-size scan on a fixed 100-query Tools-ToolRet slice (178 unique tools) over twelve subsampled registries from $N=10$ to $7{,}278$.\footnote{95\% CIs run $10$~pp on the 100-query slice; the terminal point (0.390) matches the full-set value (0.388) within 0.2~pp.} We compare Regex, Full-Ctx (Nova Micro full-context picker, 480k character budget, alphabetic truncation when exceeded), and Ours+BM25 ($k{=}15$ rerank). We show Match@1 vs.\ $N$ and decompose the pipeline into retrieval recall and reranker conditional accuracy (Figure~\ref{fig:scaling-within}).

Below $N=500$, full-context LLM picking works; above it, Full-Ctx drops $-13$~pp per doubling and collapses once alphabetic truncation triggers at $N \geq 5{,}000$ (4{,}023 of 7{,}278 tools silently dropped).
The crossover is around $N=500$ in this Nova Micro sweep and shifts with context size. Native-size AppWorld and MCP points are consistent with the pattern but do not establish a shared threshold (Figure~\ref{fig:scaling-cross}).

The scale decomposition explains the gap since the single-stage picking collapses from Match@1 $0.85$ to $0.12$ as $N$ grows, while the reranker's conditional accuracy holds steady at $0.70$–$0.87$ across all registry sizes. The pipeline therefore degrades far more gracefully ($0.81$ to $0.39$). Further gains require better retrieval (Appendix~\ref{app:scale-implications}).

\subsection{Cross-type generalization}
\label{sec:exp-crosstype}

We use one configuration (same enrichment template, retriever stack, and scorer weights) to produce every pipeline cell in Table~\ref{tab:main-mat1} with no per-type tuning.
Three choices make this possible: the enrichment template is type-agnostic (summary, action-led description, keywords, examples, regardless of target), the retrieval signal is type-blind (BM25 and the dense encoder score any text field the same way), and the reranker reads the same prompt shape (a ranked list of enriched profiles) across all three registries.
The results are consistent with registry size determining the advantage, but the benchmarks do not separate size from capability type.
Current benchmarks limit the cross-type claim. Agents-AppWorld saturates, Agents-ToolRet overlaps Tools-ToolRet, and Skills-MCP queries paraphrase capability descriptions. A stronger test requires agent and skill registries above $500$, real user intents, and multi-ground-truth trajectory labels (Appendix~\ref{app:failures}).

\subsection{Cross-model robustness}
\label{sec:exp-crossmodel}

We test four rerankers on Tools, with Nova Micro as the reference. The model ordering is stable from $n=500$ to full $n$. The ordering is Claude Sonnet 4 $>$ Nova Micro $\approx$ Nova Lite $\gg$ Claude 3.5 Haiku. At full $n$, Claude Sonnet 4 improves Match@1 by only $3$~pp, consistent with conditional accuracy inside the $0.70$--$0.87$ band (Qwen3 Next and Mistral Large 3, Table~\ref{tab:crossmodel-open}). Claude 3.5 Haiku is the exception, collapsing from 0.496 to 0.280 on output truncation. Nova Micro is the production choice; Claude Sonnet 4 is acceptable only on the smaller registry (Table~\ref{tab:crossmodel}; Appendix~\ref{app:detail-tables}).

\subsection{Efficiency}
\label{sec:exp-efficiency}

We compare retrieve-then-rank against the in-context routing baselines on accuracy and cost. Full per-row breakdowns (tokens, dollars, and latency) appear in Appendix~\ref{app:detail-tables}, Table~\ref{tab:efficiency}. All pricing uses Nova Micro managed-endpoint retail rates. Baseline rows reflect real token counts. Pipeline rows are post-hoc estimates ($\pm5$--$10\%$).
On Tools-ToolRet, Ours+BM25 is more accurate than every LLM-using in-context baseline. It is ${\sim}$\xmult{2} cheaper than Search\&Pick and ${\sim}$\xmult{70} cheaper than Full-Ctx (\$0.066 vs.\ \$4.48 per 1{,}000 queries).
On both large ToolRet rows, it is more accurate and uses fewer tokens than Full-Ctx and Search\&Pick. Full-Ctx is cheaper on Agents-AppWorld, and lexical baselines are more accurate on Skills-MCP.
Four failure examples are in Appendix~\ref{app:failures}.

\section{Deployment}
\label{sec:deployment}

The deployment evidence is architectural rather than an online evaluation, and our operational evidence is limited to cost and latency (Table~\ref{tab:efficiency}).
We deploy the pipeline in production as the capability-discovery layer for an internal multi-agent platform.
All four MATS types coexist in a single index with shared enrichment schema and scorer weights.
Offline enrichment runs at registration time. An LLM rewrites sparse metadata into the five-field profile. The profile is stored in a key-value store and encoded with Titan Embed V2. Generation cost is paid once per capability update, never per query.
The online path is a serverless AWS Lambda: BM25 retrieval over the enriched corpus, neural re-scoring via the dense index, and a single Nova~Micro reranker call over $k=15$ candidates, with all four scoring signals active where the registry includes trust and type metadata.

The production registry exceeds $500$ across all four MATS types, in the regime where single-stage LLM routing fails ($-13$~pp Match@1 per doubling past the crossover), so retrieve-then-rank is an architectural requirement, not a preference.

The public configuration uses two signals (LLM + BM25); production also uses Quality and Intent. On an internal registry with the required trust and type metadata, these signals add $+4.5$~pp Match@1 ($p=0.031$), with retrieval fixed (Appendix~\ref{app:signals}). We retain enrichment because production metadata includes the sparse conditions under which the ablation in \S\ref{sec:exp-ablations} shows a retrieval gain.

\section{Conclusion}
\label{sec:conclusion}
\vspace{-0.2em}
We recast capability discovery as information retrieval. An offline enrichment step converts sparse registry metadata into searchable profiles, and an online retrieve-then-rank pipeline returns a ranked shortlist without invoking any candidate. In-context routing's Match@1 collapses ($0.85 \to 0.12$), while retrieve-then-rank degrades gently ($0.81 \to 0.39$), with the crossover around $500$ entries in the Nova Micro sweep. At full scale the pipeline leads Search\&Pick by 6.5~pp at about half the cost. It reduces cost \xmult{70} versus Full-Ctx. The pipeline is in production.

\vspace{-0.2em}
\section*{Limitations}
Our strongest conclusions come from ToolRet ($n=7{,}278$).  The agent and skill benchmarks have construction limits (\S\ref{sec:exp-main}, \S\ref{sec:exp-ablations}) that prevent independent cross-type claims. Enrichment is neutral-to-negative on well-documented registries. Sparse metadata does not occur naturally in the evaluated benchmarks, so we reproduce it through controlled degradation. Quality and Intent require trust and type metadata absent from public registries; public numbers reflect the two-signal pipeline only. The production registry indexes models alongside the other types under the same schema, but we report no model-retrieval numbers. Existing model-routing literature addresses that case, and no public benchmark isolates model-as-capability retrieval.

\section*{Ethical considerations}
The discovery stage ranks candidates without invoking them, so it does not run candidate code. Provider-controlled metadata can bias retrieval, so operators should version profiles and monitor exposure. The public benchmarks involve no human subjects or private data.

\bibliography{main}

\appendix

\section{Detailed results}
\vspace{-0.25em}
\label{app:detail-tables}

The cost breakdown (Table~\ref{tab:efficiency}), four-reranker robustness sweep (Table~\ref{tab:crossmodel}), and raw-versus-enriched ablation (Table~\ref{tab:ablation}) support the claims in \S\ref{sec:exp-efficiency}, \S\ref{sec:exp-crossmodel}, and \S\ref{sec:exp-ablations}, respectively.

\begin{table*}[t]
\centering
\small
\setlength{\tabcolsep}{5pt}
\begin{tabular}{llrrrrrl}
\toprule
Row & Method & M@1 & Tok-in/q & Tok-out/q & \$/1k q & p50 (ms) & Source \\
\midrule
\multirow{5}{*}{Tools --- ToolRet}
 & Full-Ctx  & 0.166 & 128{,}067 & 8  & 4.4835 & 5{,}841 & real \\
 & Search\&Pick    & 0.332 & 3{,}141   & 48 & 0.1166 & 1{,}556 & real \\
 & Trial\&Err  & 0.054 & 1{,}875   & 29 & 0.0697 & 1{,}532 & real \\
 & Ours+BM25          & 0.388 & 1{,}597   & 75 & 0.0664 & 1{,}290 & est. \\
 & Ours+BGE-5f        & 0.389 & 1{,}586   & 75 & 0.0660 & 1{,}116 & est. \\
\midrule
\multirow{5}{*}{Tools --- AppWorld}
 & Full-Ctx  & 0.762 & 6{,}997 & 10 & 0.2463 & 447     & real \\
 & Search\&Pick    & 0.476 & 2{,}505 & 42 & 0.0936 & 2{,}085 & real \\
 & Trial\&Err  & 0.061 & 2{,}457 & 29 & 0.0900 & 1{,}978 & real \\
 & Ours+BM25          & 0.741 & 1{,}867 & 75 & 0.0758 & 565     & est. \\
 & Ours+BGE-5f        & 0.735 & 1{,}842 & 75 & 0.0750 & 972     & est. \\
\midrule
\multirow{5}{*}{Agents --- ToolRet}
 & Full-Ctx  & 0.148 & 122{,}672 & 7  & 4.2946 & 5{,}637 & real \\
 & Search\&Pick    & 0.341 & 3{,}172   & 47 & 0.1176 & 1{,}621 & real \\
 & Trial\&Err  & 0.045 & 1{,}747   & 26 & 0.0648 & 1{,}663 & real \\
 & Ours+BM25          & 0.418 & 2{,}061   & 75 & 0.0826 & 1{,}153 & est. \\
 & Ours+BGE-5f        & 0.412 & 2{,}232   & 75 & 0.0886 & 1{,}068 & est. \\
\midrule
\multirow{5}{*}{Agents --- AppWorld}
 & Full-Ctx  & 0.986 & 553      & 5  & 0.0201 & 360     & real \\
 & Search\&Pick    & 0.973 & 5{,}487  & 41 & 0.1978 & 1{,}946 & real \\
 & Trial\&Err  & 0.014 & 1{,}441  & 23 & 0.0536 & 1{,}792 & real \\
 & Ours+BM25          & 1.000 & 26{,}906 & 60 & 0.9501 & 1{,}543 & est. \\
 & Ours+BGE-5f        & 1.000 & 26{,}906 & 60 & 0.9501 & 1{,}628 & est. \\
\midrule
\multirow{5}{*}{Skills --- MCP (leaky)}
 & Full-Ctx  & 0.968 & 29{,}110 & 6  & 1.0197 & 1{,}195 & real \\
 & Search\&Pick    & 0.829 & 2{,}472  & 35 & 0.0915 & 1{,}492 & real \\
 & Trial\&Err  & 0.624 & 1{,}125  & 23 & 0.0426 & 1{,}212 & real \\
 & Ours+BM25          & 0.807 & 2{,}290  & 75 & 0.0906 & 141     & est. \\
 & Ours+BGE-5f        & 0.806 & 2{,}290  & 75 & 0.0906 & 176     & est. \\
\bottomrule
\end{tabular}
\caption{Retrieve-then-rank costs ${\sim}$\xmult{70} less than the full-context picker at higher Match@1. Columns report Match@1, tokens per query, dollars per 1{,}000 queries, and p50 latency. Trial-and-error rows report real measured token counts; Ours rows are post-hoc prompt-shape estimates ($\pm5$--$10\%$ of real).}
\vspace{-1em}
\label{tab:efficiency}
\end{table*}

\begin{table}[t]
\centering
\footnotesize
\setlength{\tabcolsep}{3pt}
\begin{tabular}{llcc}
\toprule
Row & Reranker & M@1 & p50\,(ms) \\
\midrule
\multirow{4}{*}{\shortstack[l]{Tools--ToolRet\\($n{=}500$)}}
 & Nova Micro       & 0.682 & 750 \\
 & Nova Lite        & 0.678 & 1{,}316 \\
 & Cl.\,3.5 Haiku   & 0.496 & 11{,}196 \\
 & Cl.\,Sonnet 4    & \textbf{0.722} & 3{,}775 \\
\midrule
\multirow{4}{*}{\shortstack[l]{Tools--ToolRet\\(full)}}
 & Nova Micro       & 0.389 & 807 \\
 & Nova Lite        & 0.395 & 1{,}453 \\
 & Cl.\,3.5 Haiku   & 0.280 & 11{,}605 \\
 & Cl.\,Sonnet 4    & \textbf{0.419} & 3{,}909 \\
\midrule
\multirow{4}{*}{\shortstack[l]{Tools--\\AppWorld}}
 & Nova Micro       & 0.782 & 553 \\
 & Nova Lite        & 0.864 & 965 \\
 & Cl.\,3.5 Haiku   & 0.503 & 11{,}628 \\
 & Cl.\,Sonnet 4    & \textbf{0.946} & 8{,}833 \\
\bottomrule
\end{tabular}
\caption{Cross-model robustness of the reranker. The large-registry gain from a stronger reranker is small, consistent with accuracy inside the $0.70$--$0.87$ band.}
\label{tab:crossmodel}
\vspace{-1.9em}
\end{table}

\begin{table}[t]
\centering
\footnotesize
\setlength{\tabcolsep}{4pt}
\begin{tabular}{llccc}
\toprule
Row & Ranker LLM & \shortstack{Search\\\&Pick} & \shortstack{Ours\\+BM25} & \shortstack{Ours\\+BGE} \\
\midrule
\multirow{3}{*}{\shortstack[l]{Tools--\\ToolRet}}
& Nova Micro   & 0.332 & 0.388 & 0.389 \\
& Qwen3 Next   & 0.368 & 0.416 & \textbf{0.419} \\
& Mistral L3   & 0.365 & 0.408 & 0.415 \\
\midrule
\multirow{3}{*}{\shortstack[l]{Agents--\\ToolRet}}
& Nova Micro   & 0.341 & 0.418 & 0.412 \\
& Qwen3 Next   & 0.385 & 0.451 & 0.452 \\
& Mistral L3   & 0.388 & 0.451 & \textbf{0.458} \\
\bottomrule
\end{tabular}
\caption{Cross-model generalization of the pipeline (not just the reranker). Match@1 at full $n$. Ours beats the strongest in-context baseline (Search\&Pick) under all LLMs: \numlist{+5.6/+5.1/+5.0}~pp on Tools-ToolRet and \numlist{+7.7/+6.7/+7.0}~pp on Agents-ToolRet. The Qwen3~Next ranker is $\sim3$~pp of Nova Micro at full $n$, so the prompt is not tuned. Bold marks the best result.}
\label{tab:crossmodel-open}
\vspace{-2em}
\end{table}

\begin{table}[t]
\centering
\small
\setlength{\tabcolsep}{4pt}
\begin{tabular}{llccc}
\toprule
Benchmark & Cond. & M@1 & M@3 & R@$k$ \\
\midrule
\multirow{2}{*}{AppWorld (332)} & Raw      & 0.782 & 0.884 & 0.420 \\
                                & Enriched & 0.782 & 0.918 & 0.400 \\
\midrule
\multirow{2}{*}{ToolRet (1{,}000)} & Raw      & 0.124 & 0.137 & 0.135 \\
                                   & Enriched & 0.125 & 0.139 & 0.138 \\
\midrule
\multirow{2}{*}{MCP Skills (859)} & Raw      & \textbf{0.929} & \textbf{0.977} & \textbf{0.988} \\
                                  & Enriched & 0.885 & 0.939 & 0.951 \\
\bottomrule
\end{tabular}
\caption{Enrichment ablation behind \S\ref{sec:exp-ablations}. LLM enrichment adds no measurable gain on well-documented public registries. Raw vs.\ LLM-enriched registry, all else held fixed. The ToolRet rows use a 1{,}000-tool subset, so absolute numbers are low but the raw-vs-enriched delta is the quantity of interest.}
\label{tab:ablation}
\end{table}

\vspace{-0.3em}
\section{Extended ranking metrics}
\vspace{-0.3em}
\label{app:full-table1}

The single-pick baselines emit one capability per query, so M@3, M@5, MRR, and R@15 are undefined. We omit them from both tables. Regex emits a deterministic ranked list (regex hits first, then alphabetic). Its lists are long enough for M@3, M@5, and MRR (Table~\ref{tab:rank-metrics}) but not for R@15, so it is absent from that table (Table~\ref{tab:r15}).
\begin{table*}[t]
\centering
\small
\setlength{\tabcolsep}{5pt}
\begin{tabular}{llcccccc}
\toprule
Row & Metric & Regex & BM25 & BGE-5f & Ours+BM25 & Ours+BGE-5f & Ours+Titan \\
\midrule
\multirow{3}{*}{Tools --- ToolRet}
 & M@3 & 0.314 & 0.423 & 0.472 & 0.493 & 0.502 & \textbf{0.523} \\
 & M@5 & 0.354 & 0.464 & 0.520 & 0.523 & 0.539 & \textbf{0.558} \\
 & MRR & 0.285 & 0.378 & 0.424 & 0.446 & 0.454 & \textbf{0.467} \\
\midrule
\multirow{3}{*}{Tools --- AppWorld}
 & M@3 & 0.687 & 0.728 & 0.721 & 0.884 & 0.898 & \textbf{0.918} \\
 & M@5 & 0.755 & 0.837 & 0.823 & 0.898 & 0.932 & \textbf{0.946} \\
 & MRR & 0.576 & 0.631 & 0.594 & 0.820 & 0.819 & \textbf{0.832} \\
\midrule
\multirow{3}{*}{Agents --- ToolRet}
 & M@3 & 0.324 & 0.445 & 0.479 & 0.523 & \textbf{0.533} & 0.519 \\
 & M@5 & 0.368 & 0.485 & 0.528 & 0.546 & \textbf{0.567} & 0.550 \\
 & MRR & 0.288 & 0.402 & 0.435 & 0.475 & \textbf{0.479} & 0.464 \\
\midrule
\multirow{3}{*}{Agents --- AppWorld}
 & M@3 & \textbf{1.000} & \textbf{1.000} & \textbf{1.000} & \textbf{1.000} & \textbf{1.000} & \textbf{1.000} \\
 & M@5 & \textbf{1.000} & \textbf{1.000} & \textbf{1.000} & \textbf{1.000} & \textbf{1.000} & \textbf{1.000} \\
 & MRR & 0.997 & \textbf{1.000} & \textbf{1.000} & \textbf{1.000} & \textbf{1.000} & \textbf{1.000} \\
\midrule
\multirow{3}{*}{Skills --- MCP}
 & M@3 & 0.811 & 0.964 & 0.958 & 0.879 & 0.880 & \textbf{0.981} \\
 & M@5 & 0.854 & 0.980 & 0.974 & 0.902 & 0.911 & \textbf{0.988} \\
 & MRR & 0.736 & 0.923 & 0.928 & 0.852 & 0.855 & \textbf{0.962} \\
\bottomrule
\end{tabular}
\caption{Ours+Titan leads on Tools-ToolRet, AppWorld, and Skills-MCP; on Agents-ToolRet, Ours+BGE-5f leads. The saturated Agents-AppWorld row ties. The single-pick baselines (Full-Ctx, Search\&Pick, and Trial\&Err) emit one capability per query, so these metrics are undefined for them and their columns are omitted.}
\label{tab:rank-metrics}
\vspace{-0.5em}
\end{table*}

\begin{table*}[t]
\centering
\small
\setlength{\tabcolsep}{6pt}
\begin{tabular}{lccccc}
\toprule
Row & \shortstack{BM25\\M@15} & \shortstack{BGE-5f\\M@15} & \shortstack{Titan\\M@15} & \shortstack{Ours+BM25\\R@15} & \shortstack{Ours+BGE-5f\\R@15} \\
\midrule
Tools --- ToolRet  & 0.549 & 0.618 & \textbf{0.625} & 0.485 & 0.517 \\
Tools --- AppWorld & 0.952 & 0.966 & \textbf{0.993} & 0.262 & 0.261 \\
Agents --- ToolRet & 0.569 & \textbf{0.626} & 0.616 & 0.508 & 0.538 \\
Agents --- AppWorld& \textbf{1.000} & \textbf{1.000} & \textbf{1.000} & \textbf{1.000} & \textbf{1.000} \\
Skills --- MCP     & 0.991 & 0.992 & \textbf{0.994} & 0.951 & 0.992 \\
\bottomrule
\end{tabular}
\caption{Retrieval at $k=15$. Standalone columns report Match@15, the query-level any-hit rate. Ours columns report Recall@15, the fraction of all ground-truth capabilities retrieved. Single-pick baselines are omitted.}
\label{tab:r15}
\vspace{-1.4em}
\end{table*}

\vspace{-0.5em}
\section{Retrieval vs. reranking}
\vspace{-0.2em}
\label{app:scale-implications}

\begin{enumerate}[nosep]
  \item \emph{Two stages prevent the single-stage collapse.} Single-stage LLM picking (Full-Ctx) collapses at scale because the LLM has to consider thousands of items at once. Our pipeline's reranker always sees exactly $k=15$. Full-Ctx goes 0.85 $\to$ 0.12 across the scan; our pipeline's reranker stays 0.85 $\to$ 0.70.
  \item \emph{The reranker is not the constraint.} ${\sim}0.87$ conditional accuracy at $N=100$ with full retrieval recall is the highest value we measure.
  \item \emph{The remaining end-to-end gains must come from retrieval.} On the scan slice at $N=7{,}278$, Match@1 is ${\sim}0.39$. A GT capability reaches the top 15 for ${\sim}56\%$ of queries, and the reranker ranks it first with ${\sim}0.70$ accuracy ($0.39 \approx 0.56 \times 0.70$). This factorization shows that retrieval limits further gains.
  \item \emph{The pipeline's advantage is largest at mid-$N$ (50--1{,}000).} $N \leq 25$ retrieval is identity and Ours $\approx$ Full-Ctx; $N \geq 5{,}000$ Ours is bottlenecked by BM25 recall, not its own design; in between, Ours leads by 7--17~pp.
  \item \emph{Why a $\geq 5$~pp BGE-over-BM25 Match@1 gain does not materialize.} At the registry sizes we test, the reranker's conditional accuracy is flat across stages, so a $+6.9$~pp Recall@15 advantage for BGE does not propagate to end-to-end Match@1. Recall@15 and end-to-end Match@3/Match@5 separate the retrievers.
\end{enumerate}

Comparisons are consistent with the sweep, but only Figure~\ref{fig:scaling-within} estimates the crossover at $N=500$.

\begin{figure}[t]
\centering
\includegraphics[width=\columnwidth]{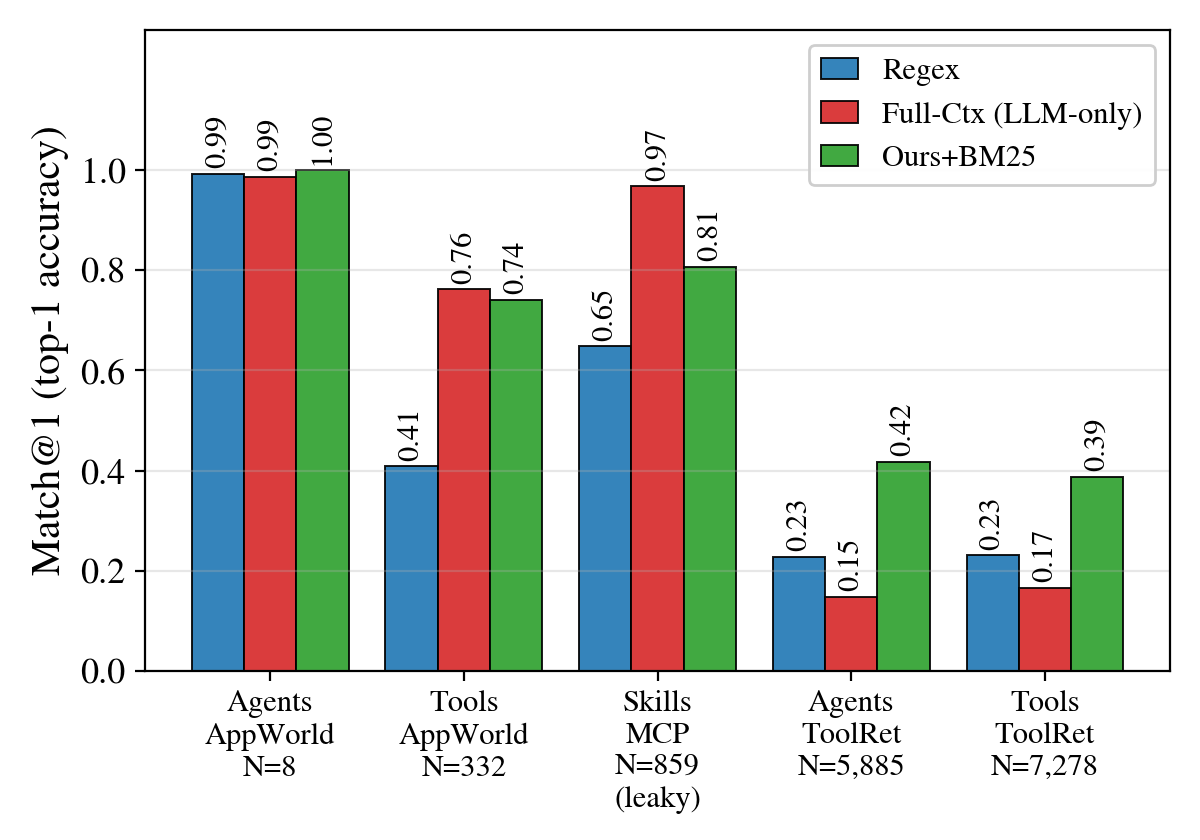}
\caption{Cross-sectional benchmark comparison. Full-Ctx wins or ties on the small rows, while Ours+BM25 leads on the large ToolRet rows. Full-Ctx uses 500-candidate subsamples on ToolRet and Skills-MCP (Table~\ref{tab:main-mat1}), so these points do not estimate a shared threshold.}
\label{fig:scaling-cross}
\end{figure}

\vspace{-0.1em}
\section{Failure modes}
\vspace{-0.3em}
\label{app:failures}

\paragraph{Category 1 -- Retrieval miss.} \textit{Tools-ToolRet.} A query asks to ``develop an emotion analysis system for customer satisfaction over the phone for a Russian telecom company.'' The ground-truth tool is a Russian speech-emotion model; the system predicts a generic customer-satisfaction scorer. The query shares no keywords with the ground-truth tool name, and BM25 misses it. This failure mode motivates the neural retriever in \S\ref{sec:exp-retrieval}.

\paragraph{Category 2 -- Picked-more-generic.} \textit{Skills-MCP.} A query asks to ``list running code-sandbox-mcp containers.'' The ground-truth tool is the named sandbox server, retrieved at rank 3; the system instead promotes a generic Docker tool. It over-weights the surrounding intent (``running containers'') despite the query naming the correct tool.

\paragraph{Category 3 -- Picked outside ground-truth set.} \textit{Tools-AppWorld.} A query asks to ``add a comment to all Venmo payments I received from coworkers and like those payments.'' The ground truth is a 9-tool trajectory; the system predicts a single tool that treats the comment and the like as one action. Match@1 penalizes this; Recall@15 does not.

\paragraph{Category 4 -- Full-Ctx miss.} \textit{Tools-ToolRet, $n{=}500$.} A query asks ``show me my history for today.'' The ground-truth tool exists in the registry but Full-Ctx returns null. At full scale, prompt truncation drops 55\% of the registry (4{,}023 of 7{,}278 tools) and creates additional misses. This truncation causes the collapse in \S\ref{sec:exp-scale}.

\section{Enrichment under sparse metadata}
\label{app:sparsity}

We reuse the 1{,}000-tool ToolRet pool from Table~\ref{tab:ablation} and vary only its input metadata. \textbf{Full} retains the upstream name and description, \textbf{hint} retains the name and first sentence (at most 150 characters), and \textbf{name-only} removes the description. For each level, the raw arm indexes that input directly and the enriched arm first applies the same enrichment procedure. Both use BM25 at $k=15$, the Nova Micro reranker, and the weights in \S\ref{sec:method-online}.

We evaluate the 1{,}252 queries whose ground-truth capability occurs in the pool. This differs from the original ablation in Table~\ref{tab:ablation}, which uses the full query set and a raw registry that already contains keywords added during preprocessing. The present comparison starts from the upstream fields and isolates how much enrichment recovers as those fields are removed.

\begin{table}[t]
\centering
\footnotesize
\setlength{\tabcolsep}{3.8pt}
\begin{tabular}{lccc}
\toprule
Input level & Raw & Enriched & R@15 raw\,$\to$\,enr. \\
\midrule
Full description   & 0.720 & 0.778 & $0.905\to0.933$ \\
First-sentence hint & 0.685 & 0.776 & $0.888\to0.929$ \\
Name-only          & 0.085 & 0.341 & $0.134\to0.467$ \\
\bottomrule
\end{tabular}
\vspace{-0.8em}
\caption{Metadata-sparsity ablation ($n=1{,}252$ per cell). Match@1 gains are $+5.8$, $+9.1$, and $+25.6$~pp from full to name-only input.}
\label{tab:sparsity}
\vspace{-0.5em}
\end{table}

\begin{figure}[t]
\centering
\includegraphics[width=\columnwidth]{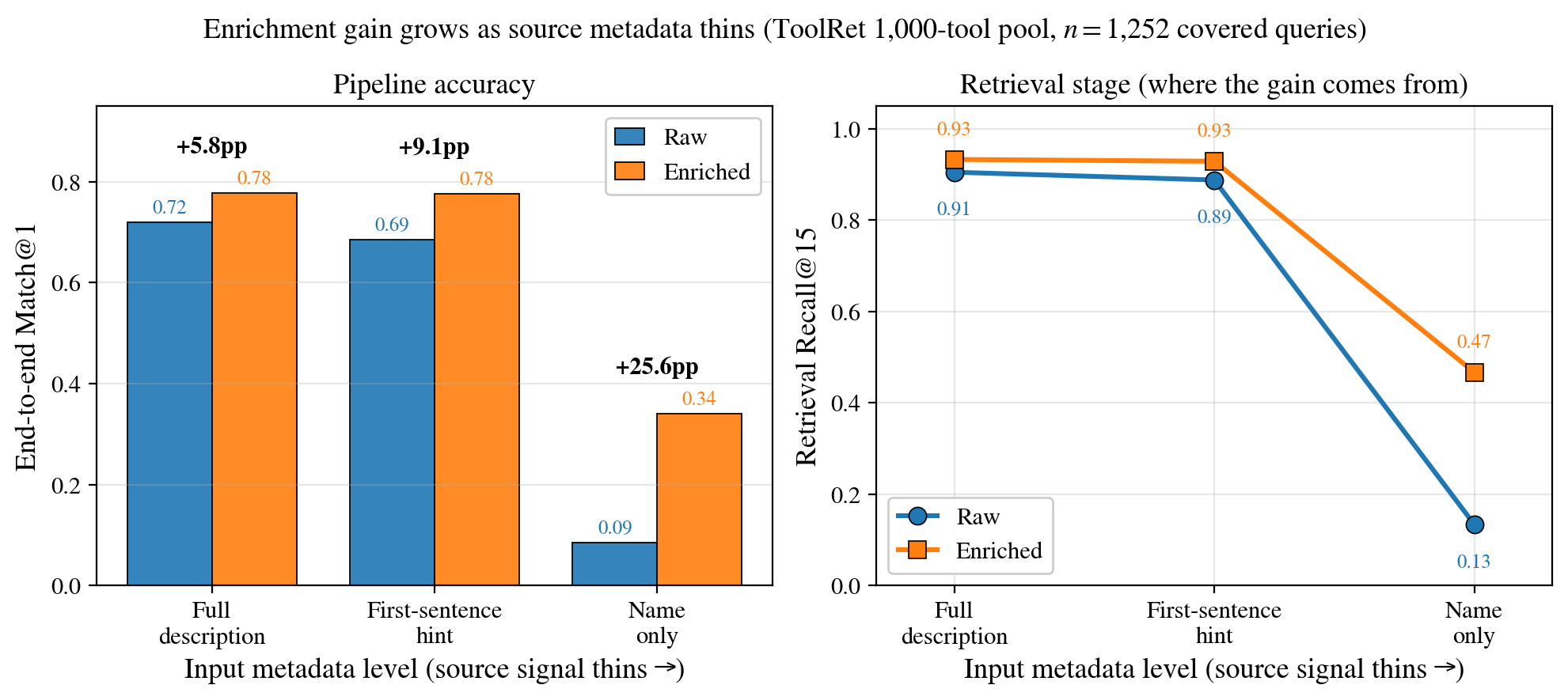}
\caption{Enrichment acts primarily through retrieval. As input metadata is removed, the raw-versus-enriched gap widens far more in Recall@15 (right) than in Match@1 (left).}
\label{fig:sparsity}
\end{figure}

At name-only, enrichment raises Recall@15 from $0.134$ to $0.467$, while rerank conditional accuracy changes from $0.63$ to $0.73$. Enriched inputs perform similarly (Match@1 $0.776$ and $0.778$), so one descriptive sentence supplies most of what this enricher uses. For 867 of 1{,}000 name-only entries, it leaves the capability summary empty rather than inferring one from an opaque name.

\section{Results by ToolRet source}
\label{app:bysource}

ToolRet combines queries from 35 upstream sources. We group the existing full-$n$ results by \texttt{metadata.subset} to test whether a single source accounts for the aggregate gain. Each group remains part of the same benchmark and is evaluated against the same registry.

\begin{table}[htbp]
\centering
\footnotesize
\setlength{\tabcolsep}{3pt}
\begin{tabular}{lrccc}
\toprule
Source & $n$ & Ours+Titan & S\&Pick & $\Delta$ \\
\midrule
\multicolumn{5}{l}{\emph{Tools --- ToolRet}}\\
toolbench            & 1{,}100 & 0.518 & 0.490 & $+0.028$ \\
apigen               & 1{,}000 & 0.702 & 0.611 & $+0.091$ \\
toolace              & 1{,}000 & 0.694 & 0.620 & $+0.074$ \\
toolink              & 497 & 0.579 & 0.348 & $+0.231$ \\
apibank              & 101 & 0.614 & 0.426 & $+0.188$ \\
gorilla-huggingface  & 500 & 0.148 & 0.180 & $-0.032$ \\
\emph{long tail (19 src)} & 798 & 0.325 & 0.246 & $+0.079$ \\
\midrule
\multicolumn{5}{l}{\emph{Agents --- ToolRet}}\\
toolace              & 1{,}000 & 0.715 & 0.637 & $+0.078$ \\
apigen               & 1{,}000 & 0.663 & 0.610 & $+0.053$ \\
toolbench            & 1{,}100 & 0.475 & 0.524 & $-0.048$ \\
craft-math-algebra   & 280 & 0.482 & 0.257 & $+0.225$ \\
apibank              & 101 & 0.624 & 0.396 & $+0.228$ \\
\bottomrule
\end{tabular}
\caption{Match@1 for representative ToolRet sources. Across all sources with $n\geq100$, Ours+Titan leads Search\&Pick on 13/16 Tools sources and 12/16 Agents sources. ``Long tail'' pools 19 smaller Tools sources.}
\label{tab:bysource}
\end{table}

\paragraph{A broken source.} The \texttt{ultratool} source (500 queries; 6.3\% of ToolRet) is malformed. Every label embeds the same tool document, so 287 referenced ground-truth tools are absent from the registry, and all methods score near zero on it. Excluding it raises every method's Tools Match@1 by ${\sim}2$~pp and leaves the ordering unchanged.

\section{Tool-tuned retriever}
\label{app:tooltuned}

We test ToolBench-IR, the BERT-base retrieval model released with ToolBench and fine-tuned on tool-API relevance \citep{qin2024toolllm}, as the first stage on both large rows. All retrievers use the same five-field documents and $k=15$.

\begin{table}[htbp]
\vspace{0.1em}
\centering
\footnotesize
\setlength{\tabcolsep}{4pt}
\begin{tabular}{llccc}
\toprule
Row & Retriever & Standalone & Standalone & E2E \\
    &           & M@1 & R@15 & M@1 \\
\midrule
\multirow{3}{*}{Tools}
 & BM25            & 0.311 & 0.549 & 0.388 \\
 & ToolBench-IR    & 0.354 & 0.608 & 0.378 \\
 & Titan V2        & 0.358 & \textbf{0.625} & \textbf{0.397} \\
\midrule
\multirow{3}{*}{Agents}
 & BM25            & 0.336 & 0.569 & \textbf{0.418} \\
 & ToolBench-IR    & \textbf{0.362} & 0.596 & 0.381 \\
 & BGE-large 5f    & 0.360 & \textbf{0.626} & 0.412 \\
\bottomrule
\end{tabular}
\caption{Tool-tuned and general retrievers on the two full ToolRet query sets ($n=7{,}961$). ToolBench-IR improves standalone Match@1 over BM25 but does not exceed the best general encoder on Recall@15 or end-to-end (E2E) Match@1.}
\label{tab:tooltuned}
\end{table}

\section{Contribution of quality and intent}
\label{app:signals}

Public benchmarks lack the trust and type metadata that Quality and Intent need, so these two signals drop out (\S\ref{sec:method-online}). We compare the two- and four-signal configurations on an internal single-turn agent benchmark ($n=221$) whose registry includes both fields. We hold retrieval fixed and report only paired performance differences.

\begin{table}[htbp]
\vspace{2.4em}
\centering
\footnotesize
\setlength{\tabcolsep}{5pt}
\begin{tabular}{lc}
\toprule
Metric & Four-signal minus two-signal \\
\midrule
Match@1   & $+4.5$~pp \\
Match@3   & $+5.0$~pp \\
Recall@15 & $0.0$~pp \\
\bottomrule
\end{tabular}
\caption{Paired effect of adding Quality and Intent on the internal benchmark. Fourteen queries change from miss to hit at rank~1 and four change from hit to miss (exact McNemar $p=0.031$). Recall is unchanged because retrieval is shared.}
\label{tab:signals}
\end{table}

\end{document}